\documentclass[twocolumn]{revtex4}

\usepackage[english]{babel}

\usepackage{amsmath}
\usepackage{graphicx}
\usepackage[colorlinks=true, allcolors=blue]{hyperref}

\begin{document}

\title{Dependence of Equilibrium Propagation Training Success on Network Architecture}

\author{Qingshan Wang}
\affiliation{Max Planck Institute for the Science of Light, Staudtstraße 2, 91058 Erlangen, Germany}
	
\author{Clara C. Wanjura}
\affiliation{Max Planck Institute for the Science of Light, Staudtstraße 2, 91058 Erlangen, Germany}
	 
\author{Florian Marquardt}
\affiliation{Max Planck Institute for the Science of Light, Staudtstraße 2, 91058 Erlangen, Germany}
\affiliation{University of Erlangen-Nuremberg, 91058 Erlangen, Germany}
	
\date{\today}

\begin{abstract}
The rapid rise of artificial intelligence has led to an unsustainable growth in energy consumption. This has motivated progress in neuromorphic computing and physics-based training of learning machines as alternatives to digital neural networks. Many theoretical studies focus on simple architectures like all-to-all or densely connected layered networks. However, these may be challenging to realize experimentally, e.g. due to connectivity constraints. In this work, we investigate the performance of the widespread physics-based training method of equilibrium propagation for more realistic architectural choices, specifically, locally connected lattices. We train an XY model and explore the influence of architecture on various benchmark tasks, tracking the evolution of spatially distributed responses and couplings during training. Our results show that sparse networks with only local connections can achieve performance comparable to dense networks. Our findings provide guidelines for further scaling up architectures based on equilibrium propagation in realistic settings.
\end{abstract}

\maketitle

\section{Introduction}

In the last decade, artificial neural networks (ANN) have proven their capability in addressing complex tasks that are out of the reach of conventional algorithms. However, along with the astonishing achievements of ANNs, the amount of energy consumed to support the computational demands required by modern artificial intelligence (AI) applications has grown rapidly. This motivates the study of neuromorphic computing, where one explores the training of physical systems that could replace conventional digital ANNs. Specifically, in neuromorphic computing one views physical systems as information-processing machines, considers some of their degrees of freedom as input and output, and updates trainable parameters to achieve the desired input-output mapping. For each neuromorphic platform, one has to figure out suitable methods to do inference and training. 

Many examples of neuromorphic systems have by now been explored. Even early developments in machine learning like the Hopfield network \cite{hopfield1982neural} and the Boltzmann machine \cite{ackley1985learning} already had a direct connection to (statistical) physics and were even experimentally implemented \cite{farhat1985optical, ticknor1987optical, kim2009highly}.
Nowadays, neuromorphic platforms more generally range from optical systems \cite{wetzstein2020inference} to solid state systems like spintronics or arrays of memristors \cite{li2018review}, and from coupled oscillators \cite{hoppensteadt1999oscillatory} to mechanical networks \cite{dillavou2022demonstration}.


However, training neuromorphic systems is challenging.
Training a physical network is fundamentally different from training conventional ANNs, which can simply rely on the mathematical algorithm of backpropagation. 
Physical neural networks either have to be trained in simulation or the training has to rely on physics-based training methods which only exist for certain classes of systems.



In the past few years, several techniques for the efficient training of neuromorphic devices have been proposed, explored, and partially already implemented (for a review see \cite{momeni2025training}). Examples include physics-aware training \cite{wright2022deep} (a hybrid technique combining backpropagation in simulation with physical inference), forward-forward training which optimizes layer activations locally \cite{oguz2023forward,momeni2023backpropagation}, variations of contrastive learning (e.g. \cite{de2025learning}), and a variety of approaches addressing optical neural networks: backpropagation through the linear components \cite{pai2023experimentally}, training updates via physical backpropagation relying on particular nonlinearities \cite{wagner1987multilayer,spall2025training}, gradients extracted via scattering in situations where nonlinear processing is achieved in linear setups \cite{wanjura2024fully}, a recently introduced general approach for scattering-based training in arbitrary nonlinear optical setups \cite{cin2025training}, and Hamiltonian Echo Backpropagation with training updates based on physical dynamics \cite{lopez2023self}.

While optical devices operate far out of equilibrium, there is also the possibility to exploit interacting equilibrium systems as a neuromorphic platform. In this domain, one observes the system after equilibration under constraints given by the current input. For this wide range of systems, i.e. energy-based models, there is a powerful approach for the physics-based extraction of training gradients, namely Equilibrium Propagation (EP) \cite{scellier2017equilibrium}.

Over time, several variations of EP have been proposed and implemented, such as coupled learning \cite{stern2021supervised}.  Various extensions, e.g. to vector fields \cite{scellier2018generalization}, to Lagrangian systems \cite{pourcel2025lagrangian, massar2025equilibrium}, and for more accurate gradients \cite{ernoult2020equilibrium, laborieux2021scaling}, have been presented. Furthermore, autonomous parameter updates driven by physical interactions have been explored \cite{falk2025temporal, ernoult2020equilibrium, scellier2022agnostic}. Recently, EP has even been extended to quantum many-body systems \cite{wanjura2025quantum, scellier2024quantum, massar2025quantum}.

In terms of applications to possible physical scenarios, up to now, EP and its extensions have been explored numerically for Hopfield-like models \cite{scellier2017equilibrium}, linear electrical resistor networks \cite{stern2021supervised}, nonlinear resistive networks \cite{kendall2020training}, and coupled oscillators \cite{wang2024training, rageau2025training}. In experiments, EP has successfully been applied to linear resistive networks \cite{dillavou2022demonstration,wycoff2022desynchronous,stern2024training}, but other platforms such as Ising models \cite{laydevant2024training} have been considered as well.

In the present paper, we investigate the effect of network architecture on the expressivity and performance of networks trained with EP. This is crucial, since physical implementations often operate under constraints such as local connectivity, different from the convenient theoretical assumption of all-to-all connectivity. In particular, we focus on the XY model as a prototypical nonlinear model that can be trained using EP \cite{wang2024training}. We expect many of our qualitative results to generalize to other models. Our analysis thereby contributes to the important issue of scaling up EP-based architectures in realistic settings.

We focus on the performance and trainability of lattices with local connectivity and of stacked lattices with local inter-layer connections. These two types of architecture are of a general type and experimentally feasible. 
In this article, we first analyze the evolution of network response and trainable couplings, by examining training results on the XOR task, which serves as a standard benchmark for assessing a network’s ability to perform nonlinear computations. We then investigate the effect of system size using the Iris dataset and compare the outcomes with those obtained from training all-to-all connected networks and layered networks with dense connectivity. Notably, in contrast to all-to-all and dense-layer architectures, a lattice contains significantly fewer couplings, all of which are local. This comparison thus reveals the extent to which couplings can be reduced while preserving the functional capacity of the network. Finally, the role of architectural choices of the hidden layers in layered networks is studied by comparing the performance of different networks on full-size MNIST. We consider stacked lattices with local inter-layer connections and test the effect of weight sharing, number of channels, as well as intra-layer connections.

\section{Brief Review of Equilibrium Propagation}

    In this section, we briefly review equilibrium propagation (EP), a method for extracting approximate parameter gradients of a cost function in physical networks. EP assumes a physical network modeled as an energy-based model (EBM) in the context of supervised learning. The network energy is denoted by $E(\{x\}, \theta)$, where $\{x\}$ represents the network configuration and $\theta$ the trainable parameters. In EP, $E$ is referred to as the \emph{internal energy function}.
    
    We consider a system composed of multiple nodes. Some nodes are designated as \emph{input nodes}, whose states are fixed during evolution to the steady state, while others are \emph{output nodes}, whose steady-state configurations define the network output. Supervised learning requires an additional energy function $C(x_\mathrm{out}, y^\tau)$ describing the interaction between the output and the target $y^\tau$. The total energy is then
    
    \begin{equation}
        F(x; \theta) = E(x, \theta) + \beta C(x_\mathrm{out}, y^\tau),
    \end{equation}
    in which $\beta$ controls the coupling strength. The steady state of the system is obtained by solving
    
    \begin{equation}
        \dot{x}_i = -\frac{\partial F}{\partial x_i} = -\frac{\partial E}{\partial x_i} - \beta \frac{\partial C}{\partial x_i}(x_\mathrm{out}, y^\tau).
    \end{equation}
    
    EP proceeds in two phases: the \emph{free phase} and the \emph{nudged phase}. In the free phase, the input nodes are fixed and the external energy is disabled ($\beta=0$), yielding the free equilibrium $x_0^*$. In the nudged phase, $\beta$ is set to a small value to slightly perturb the network toward the desired output, producing the \emph{nudged equilibrium} $x_\beta^*$. The parameter gradient is then approximated by the difference between the two steady states:
    
    \begin{equation}
        \begin{split}
            L(\theta, y) &= C(x^*_{0, \mathrm{out}}(\theta), y), \\
            \frac{\partial L}{\partial \theta} &= \left. \frac{d}{d \beta} \right|_{\beta=0} \frac{\partial E}{\partial \theta}(x^*_\beta, \theta) \\
            &= \lim_{\beta \rightarrow 0} \frac{1}{\beta} \left( \frac{\partial E}{\partial \theta}(x^*_\beta, \theta) - \frac{\partial E}{\partial \theta}(x^*_0, \theta) \right).
        \end{split}
    \end{equation}

    Generally, $\frac{\partial E}{\partial \theta}$ is a local function of a small number of nodes \cite{scellier2017equilibrium, wang2024training}, enabling parameter updates based on knowledge from local configurations.
    
    An alternative way to introduce the nudging force is to perturb the system with a constant force given by the error signal at the free equilibrium, which we refer to as the \emph{tangent method} in this paper. Specifically, during the nudged phase, the steady state is obtained from
    
    \begin{equation}
        \dot{x}_i = -\frac{\partial E}{\partial x_i} - \beta \frac{\partial C}{\partial x_i}(x^*_0, y^\tau).
    \end{equation}
    
    Several improvements have been proposed. For instance, symmetric nudging can enhance stability \cite{laborieux2021scaling}. Methods for autonomous parameter updates have also been developed, typically by separating the network into fast and slow subsystems, where inference occurs via fast dynamics and parameter updates via slow dynamics \cite{scellier2022agnostic, ernoult2020equilibrium, falk2025temporal}.
    
    Most concrete examples of EP have focused on Hopfield-like networks, such as nonlinear resistive networks~\cite{scellier2017equilibrium, kendall2020training, laborieux2021scaling, dillavou2022demonstration}. 
    
    %
    %
    EP has also been successfully applied to more complex physical models, including Ising models \cite{laydevant2024training} and coupled oscillators \cite{wang2024training, rageau2025training}. For networks exhibiting multistability, the gradient can be estimated by averaging contributions from all possible steady states \cite{wang2024training}.

\section{Results}


    In this paper, we investigate the XY model across different architectures, employing the training method introduced in \cite{wang2024training} to address the issue of multistability. 


    \subsection{Model and Training}

    We focus on XY models of different coupling architectures. The definition of internal energy $E$ and external energy $C$ are the same as in \cite{wang2024training}:

    \begin{equation}
        E = -\sum_{<i,j>} W_{ij} \cos(\phi_i - \phi_j) - \sum_i h_i \cos(\phi_i - \psi_i)
    \end{equation}
    and
    \begin{equation}
        C = \sum_{i \in O} -\log(1+\cos(\phi_i - \phi^\tau_i))
    \end{equation}
    in which $O$ refers to all the output nodes, $\langle i,j\rangle$ refers to a coupled pair of nodes, and $h_i$ and $\psi_i$ refer to the amplitude and direction of a local bias field. During the training, the steady states are obtained by solving the following ODE with $\beta=0$ in the free phase and $\beta \neq 0$ in the nudged phase: 

    \begin{equation}
        \dot{\phi}_i = - \frac{\partial F}{\partial \phi_i} = -\frac{\partial E}{\partial \phi_i } - \beta \frac{\partial C}{\partial \phi_i}
    \end{equation}



    Training is performed using equilibrium propagation \cite{scellier2017equilibrium, wang2024training}.  
    For each architecture, we initialize the lattice with 100 different initial sets of couplings and with zero bias fields. During the training, the performance is assessed with the following distance function, again following \cite{wang2024training}:
    
    \begin{equation}
        D(\phi, \phi^\tau) = \frac{1}{4} \sum |\boldsymbol{s}_i - \boldsymbol{s}^\tau_i|^2 = \frac{1}{2} \sum 1 - \cos(\phi_i - \phi^\tau_i)
    \end{equation}

    


    Throughout all numerical experiments, we set $\beta = 0.1$.

    \subsection{Evolution of the Network Response}



    Since we are most interested in locally connected architectures, it is of particular importance to understand the response of such architectures to local perturbations, since this determines the transport of information and, as we will see, eventually the success of training. In this section, we therefore study the network response; for illustration, we chose the particular task of training the XOR function. We find that the evolution of the network structure can be described in terms of both response and coupling strength, which collectively reveal a division of the system into an affected region, sensitive to input perturbations, and a marginal region, largely inactive.  

\begin{figure}
        \centering
        \includegraphics[width=1\linewidth]{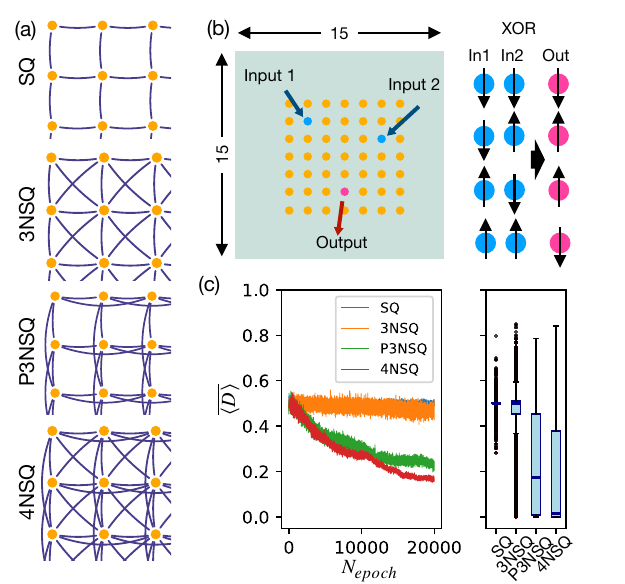}
        \caption{Lattice architectures and training performance for XOR: (a) Schematic descriptions of lattice architectures studied in this work and encoding of XOR. (b) Input-output node configuration. (c) Left: performance of lattices in training with EP, measured by the average distance function. Right: Distribution of distance at the last 100 epochs. We denote averages over input–output pairs by $\langle * \rangle$ and averages over different training cases by $\overline{*}$. } 
        \label{fig:XOR_schematic}
    \end{figure}

    We introduce the four different lattice architectures shown in Fig.~\ref{fig:XOR_schematic}~(a) that differ in their connectivity. As shown in Fig.~\ref{fig:XOR_schematic}~(b), the input and output nodes are placed near the central region of a $15 \times 15$ lattice with no direct connections between them.
    Specifically, in our numerical experiments, the input nodes are located at $(6,6)$ and $(7,10)$, while the output node is at $(10,8)$, where $(x,y)$ denotes the node in the $x$-th row and $y$-th column. The encoding protocol of true and false is shown on the right of Fig.~\ref{fig:XOR_schematic}~(b). The learning rate is chosen as $\eta = 0.01$. To mitigate the effect of multistability, networks are randomly initialized and run five times per epoch for each data sample ($M_{\text{init}} = 5$).  


    We found that for the given architecture trained on the XOR task, training with an additional nudging force at the free equilibrium (as discussed in the appendix of Ref. \cite{wang2024training}), which is slightly different from standard EP at finite $\beta$, results in a higher convergence probability and less fluctuations in the loss function. 

    \begin{equation}
        F_{i, \mathrm{nudged}, \mathrm{XOR}} = - \frac{\partial C}{\partial \phi_i}(\phi^\mathrm{free}_\mathrm{out}) = - \frac{\sin(\phi_{i,\mathrm{out}}^\mathrm{free} - \phi^\tau)}{1 + \cos(\phi_{i,\mathrm{out}}^\mathrm{free} - \phi^\tau)}
    \end{equation}
This approach increases the probability of converging to the correct result.

    The performance of the four lattice architectures, measured as the distance to the target output averaged over both data and initializations, is presented in Fig.~\ref{fig:XOR_schematic}~(c). The SQ and 3NSQ lattices show very limited ability to learn XOR, as their average distance decreases only slightly during training. Moreover, the distance distributions for these architectures are broad, indicating unstable training, where the distance fluctuates strongly during training. In contrast, the P3NSQ and 4NSQ lattices successfully learn XOR, with the 4NSQ achieving lower median error and a more concentrated distribution. We attribute this difference to the longer-range response of architectures with higher connectivity, such as P3NSQ and 4NSQ (discussed below).

\begin{figure}
        \centering
        \includegraphics[width=1.0\linewidth]{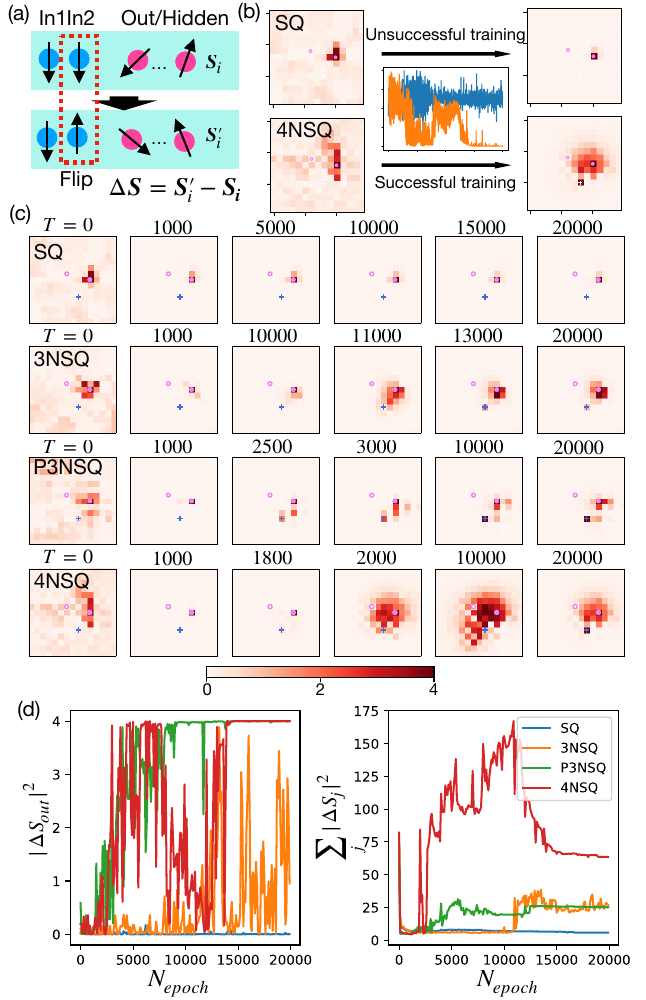}
        \caption{ Evolution of lattice response to flip on input nodes: (a) Schematic demonstration of the experiment. Nodes ``In1'' and ``In2'' are the two input nodes on the lattice. The value of In2 is flipped and the response of spins at other nodes are recorded. (b) Sample result of the experiment for the lattice response of the SQ and 4NSQ lattices. The two panels on the left show the lattice response to the flip on node In2 for the SQ and 4NSQ lattices before the training. The two panels on the right show the response after training. The panel at the center shows the evolution of the mean error over all the 4 input-output pairs for  SQ (blue line) and 4NSQ (orange line), in a typical example. (c) The evolution of lattice response for different lattices at specific stage of training. (d) The panel on the left shows the evolution of response at the output node. The panel on the right shows the sum of response at all the nodes in the lattice, as an indicator of total intensity of response.}
        \label{fig:XOR_response}
    \end{figure}

    To investigate the effect of the architecture on the evolution of the network response, we flip the second input node in the original input configuration $\downarrow,\downarrow$ and examine the resulting network behavior, as illustrated in Fig.~\ref{fig:XOR_response}~(a). This perturbation is finite and thus not amenable to linear response theory. To account for the possible presence of multistability (which may lead to different responses in different runs), we choose to characterize the response using the squared average difference. Specifically, we compute the node response as:

    \begin{equation}
        \langle | \Delta S_j  |^2 \rangle_\mathrm{steady \ states} =\langle |S_{j, \downarrow \downarrow} - S_{j, \downarrow \uparrow} \rangle_\mathrm{steady \ states} 
    \end{equation}
    which ranges from 0 to 4. To describe the overall response of any given lattice, we sum over the whole lattice, taking $\sum_j \langle | \Delta S_j |^2 \rangle_\mathrm{steady \ states}$ as an indicator.

    We analyze the evolution of the network response using representative cases. For P3NSQ and 4NSQ lattices, we select successful cases, which are more typical for these architectures, while for SQ and 3NSQ lattices, we examine unsuccessful cases, reflecting their common behavior. Selected lattices are perturbed at different training stages, and the responses are recorded. Panel (b) of Fig.~\ref{fig:XOR_response} compares typical examples of SQ and 4NSQ lattices. For the SQ lattice, training fails (blue line) and long-range responses to the perturbed input are suppressed. In contrast, a successfully trained 4NSQ lattice (orange line) evolves from a random pattern into a lattice with an affected region, closely correlated to the input flip, and a marginal region with lower response (panel (c) of Fig.~\ref{fig:XOR_response}).  

    Across all architectures, significant localization and loss of long-range response occurs during the first 1000 epochs, which can be attributed to the growth of a spatially random bias field that  suppresses longer range responses. SQ lattices fail to recover long-range responses and continue to show minimal (and localized) response to input perturbations, whereas the other three architectures recover varying degrees of longer-range responses during training. Notably, the 4NSQ lattice initially exhibits a strong, widespread response to the perturbation, which is later refined by suppressing unnecessary responses, indicating self-organization during training.  

    This behavior is also evident in the right panel of Fig.~\ref{fig:XOR_response}(d), which shows the evolution of the total response to the flip. For 4NSQ, a clear transient peak of total response emerges during training, which we interpret as the lattice dynamically identifying ``paths'' to transmit the input perturbation to the output during this stage. Ultimately, an affected region forms between input and output nodes, while a marginal region exhibits minimal response.

    \subsection{Evolution of Network Structure}

    The evolution of network structure can be characterized by changes in the coupling strengths. Fig.~\ref{fig:XOR_response}(a) compares typical training cases for SQ and 4NSQ lattices. After 20,000 epochs, the 4NSQ lattice exhibits a pronounced strengthening of couplings within the region connecting the input and output nodes (highlighted by the red dashed circle), whereas the SQ lattice shows minimal change. Notably, this region coincides with the affected region observed in the lattice response.  

    Fig.~\ref{fig:XOR_response}(b) presents the evolution of all four lattice architectures over 20,000 epochs. In the SQ lattice, couplings remain largely unchanged, ultimately leaving the output node isolated. By contrast, the other three architectures develop substantially stronger couplings within their affected regions. Interestingly, although couplings in the 3NSQ lattice also grow significantly, it nevertheless fails to learn XOR, underscoring the crucial role of skip connections in facilitating successful training. In the broader context of artificial neural networks, skip connections are known to smoothen the loss landscape, thereby improving trainability \cite{he2016deep}.
    
    Overall, the XOR training results demonstrate both the varying learning capacities of different lattices and the influence of the architecture on performance. Lattices with skip connections—and thus higher connectivity—consistently achieve superior results. The effect is much larger than could be guessed based on the slightly larger---but still local---connectivity. 

    \begin{figure}[!htb]
        \centering
        \includegraphics[width=1.0\linewidth]{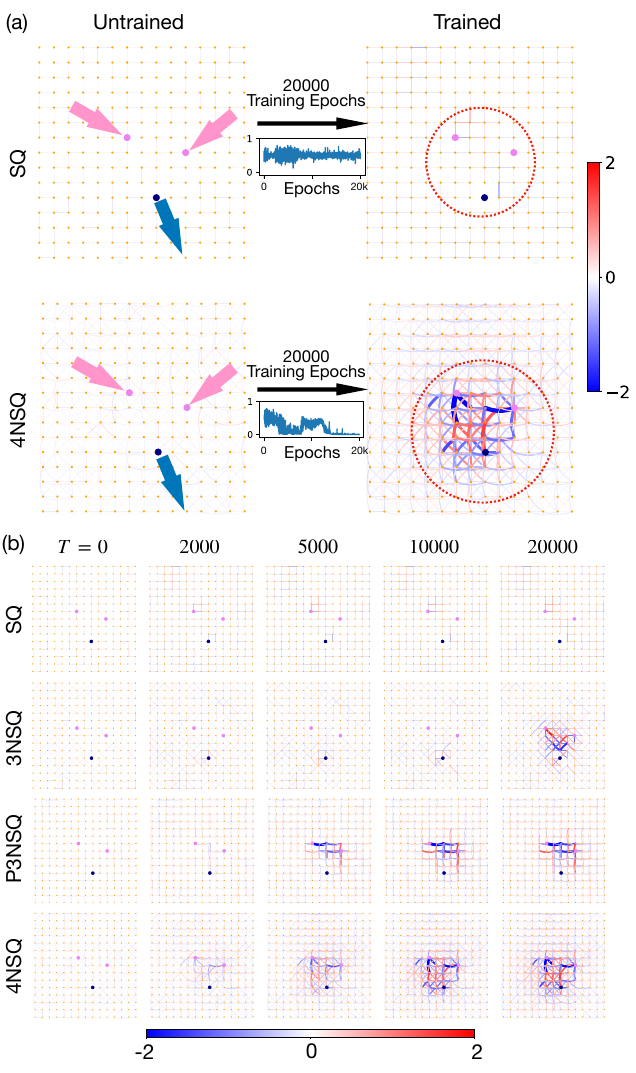}
        \caption{ Evolution of lattice response and couplings: (a) Comparison between typical training cases. On the left the connections in the SQ and 4NSQ lattices are shown. On the right the connections after training are shown. A region in which a significant change is observed is encircled. (b) The couplings of lattices of different architectures at different stages of training. }
        \label{fig:XOR_weights}
    \end{figure}

\subsection{Effect of the Lattice Size}



    We now consider the rectangular lattices of the same transverse length and examine the effect of their lateral extent which we call ``depth'' in the rest of this paper. Just as in standard artificial neural networks, it is not immediately obvious to which extent this can improve network performance, since two effects may counteract each other: the increased number of trainable parameters and flexibility vs. the problem of transporting information over larger distances (or more layers in the case of standard neural networks).
    
\begin{figure}
        \centering
        \includegraphics[width=1.0\linewidth]{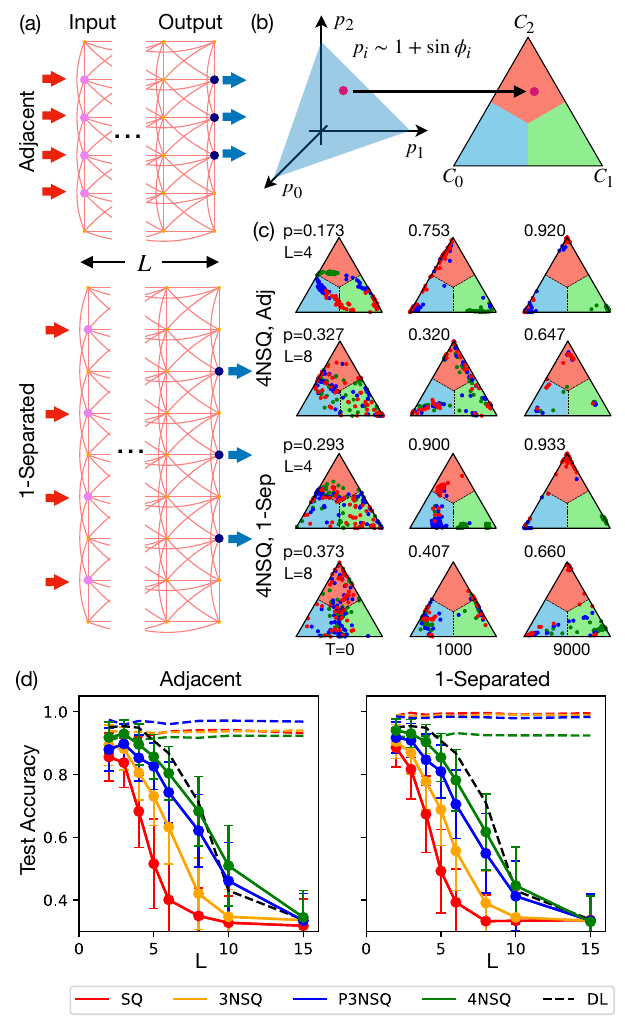}
        \caption{ Effects of network depth: (a) Schematic graph of the configuration of the tested lattices, taking 4NSQ as an example. (b) The interpretation of output configurations and their projection onto the triangular graphs. (c) The result of classification of training samples at different stages of training for lattices of different architecture and size. (d) Comparison between performance of lattices (solid lines) to all-to-all networks of the same number of parameters (dashed lines of the same colors) and densely coupled layered network (black dashed line). }
        \label{fig:Iris}
    \end{figure}

    We choose to examine the effects of network depth using the Iris dataset, which contains 50 samples for each of the three classes. The architectures employed for classification are illustrated in Fig.~\ref{fig:Iris}(a). We use the same lattice architectures as in the XOR setting. For each class, 40 samples are randomly selected for training and the remaining 10 for testing, yielding 120 training samples and 30 test samples in total. During training, the network is evaluated at specific stages by classifying data from mini-batches of 30 randomly chosen data points.  

    The learning rate is set to $\eta = 0.001$ and the time of initialization is $M_{init}=5$.  In this case (and in contrast to XOR), we find the standard EP approach to work well, i.e. the nudging force corresponds to the gradient of the external energy evaluated at the current configuration of the field (not at the given free configuration):

    \begin{equation}
        F_{i, \mathrm{nudge}} = - \frac{\partial C}{\partial \phi_i}(\phi_{\mathrm{out}}) = - \frac{\sin(\phi_{i,\mathrm{out}} - \phi^\tau)}{1 + \cos(\phi_{i,\mathrm{out}} - \phi^\tau)}
    \end{equation}

    We consider the performance of two types of lattice architectures of variable depth.  We distinguish the ``\emph{adjacent}'' configuration (upper part of panel (a)), where we place the input nodes as four consecutive nodes along one side of the lattice, from the  ``\emph{1-separated}'' configuration, where the input nodes are separated by one site each. In both cases, the output nodes are located at the opposite end of the lattice. The output node field values are interpreted as probability distributions over the three classes. These lattices can also be viewed as layered networks, where the depth is denoted by $L$.  

   We compare the performance of those architectures with that of all-to-all networks containing the same number of parameters and dense-layer (DL) networks with identical numbers of layers and nodes.

    



    The three classes of the Iris dataset are denoted as $C_0$, $C_1$, and $C_2$, with corresponding probabilities $p_0$, $p_1$, and $p_2$, where $p_i \sim 1 + \sin \phi_i$. Normalization of probability requires 
    \[
    p_0 + p_1 + p_2 = 1,
    \]  
    so any point $(p_0, p_1, p_2)$ lies within the equilateral triangle defined by the vertices $(1,0,0)$, $(0,1,0)$, and $(0,0,1)$. The result of inference for each point is the class with the highest probability. When the classes are color-coded, the triangle is partitioned into three identical regions, as illustrated in panel (b) of Fig.~\ref{fig:Iris}.  

    Panel (c) of Fig.~\ref{fig:Iris} shows the training performance for representative cases. The color of each point indicates its true class label: for instance, a red point in the red region corresponds to a correct classification of $C_2$, whereas a red point in the blue region indicates a misclassification as $C_0$. We train 4NSQ lattices with both adjacent and 1-separated architectures at varying depths $L$. In all four cases, we show the result of classification on different stages of training ($T$ refers to the time of iteration) and, as expected, training produces a systematic drift of points toward the edges and vertices of the simplex, reflecting the effect of learning.

    We investigate the effect of depth on the performance of 4NSQ lattices with both adjacent and 1-separated architectures using the Iris dataset. For shallow networks, both designs achieve training accuracies above 90\%, with the 1-separated lattice performing slightly better. At larger depth ($L=8$), performance deteriorates for both, though the 1-separated design exhibits greater stability.  

    To place these results in context, we compare lattice architectures with all-to-all networks of equal parameter count and dense-layer (DL) networks of the same depth, as shown in panel (d) of Fig.~\ref{fig:Iris}. For each configuration, 100 networks with different initial parameters are trained for 10,000 iterations (2,500 epochs), and the average test error over the last 2,000 iterations is plotted together with its variance. All-to-all networks of the same number of parameters are represented by dashed lines of the same color, while the black dashed line corresponds to a DL of equal depth (e.g., an adjacent lattice with $L=8$ corresponds to an 8-layer DL, each layer containing six nodes).  

    At small depth, all lattices perform well: 100\% test accuracy is reached in the best cases for $L \leq 3$. With increasing depth, P3NSQ and 4NSQ lattices sustain higher accuracy, with best cases attaining 100\% for $L \leq 6$ and $L \leq 8$, respectively. Beyond this range, performance gradually declines, whereas all-to-all networks maintain excellent accuracy. Nevertheless, relative to DLs of the same depth, lattices show only minor deficits. Remarkably, for $L > 10$, 4NSQ lattices outperform DLs, highlighting their potential. We find that for small depth, local connections are sufficient for good performance. For larger depth, skip connections can help to preserve information across depth, akin to residual network. 

\subsection{Networks of Layered Lattices}

    In the previous sections, we analyzed how architectural design influences network response and connection evolution, and we studied the effect of system size.

    Having established the role of architecture in ``flat" networks, we now turn to layered networks, similar to the usual arrangement in artificial neural networks. Such networks are more complex and are expected to handle more challenging tasks.  We analyze different layer architectures, showing their performance on learning full-size MNIST. Since this involves an image-classification task, layered networks with local connectivity, similar to convolutional neural networks, are suitable.

\begin{figure}[!htp]
        \centering
        \includegraphics[width=1.0\linewidth]{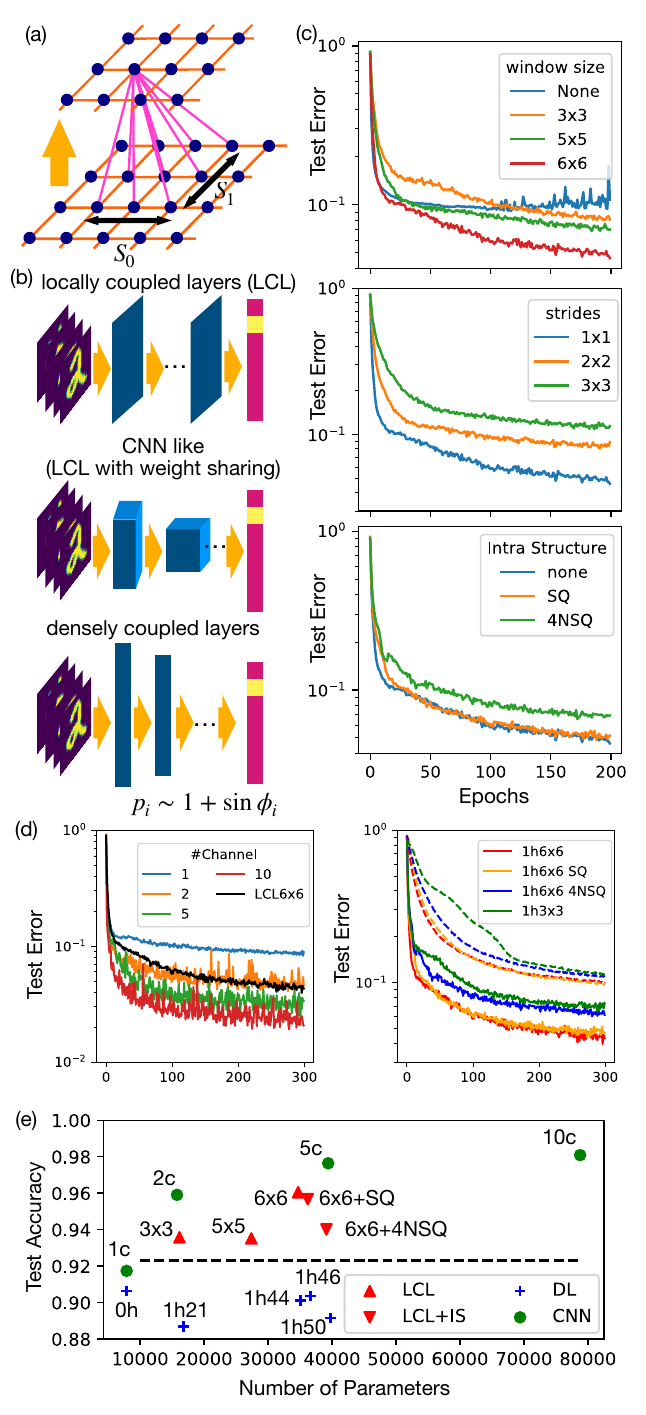}
       
        \caption{Architecture and performance of layered networks trained on MNIST, all with a single hidden layer.  
        (a) Schematic of a locally coupled lattice (LCL) layer.  
        (b) Schematic of LCL, CNN-like, and dense (DL) networks.  
        (c) LCL network performance – top: effect of window size; middle: effect of strides; bottom: effect of intra-layer couplings.  
        (d) Comparison of architectures: left, CNN-like networks with multiple channels versus LCL network with $6\times6$ window; right, LCL (solid lines) versus DL (dashed lines), with same-color lines indicating similar parameter counts.  
        (e) 
        Effect of parameter count on test accuracy. 
        For LCL and LCL+IS models, labels of the form $m \times n$ indicate the window size, 
        with the term following ``+'' denoting the intra-layer architecture. 
        In CNN-like models, the filter size is fixed at $6 \times 6$, 
        and $m$c indicates $m$ channels. 
        In DL models, labels such as ``1h21'' represent one hidden layer with 21 nodes. 
        The black dashed line indicates 92.3\% test accuracy attained by a linear classifier.}
        \label{fig:MNIST}
    \end{figure}

    A straightforward approach to constructing layered neuromorphic networks from lattices is to stack them and introduce local inter-layer couplings. Specifically, each node in an upper layer is connected to a small square array of nodes in the preceding layer, as illustrated in panel (a) of Fig.~\ref{fig:MNIST}. The spatial extent of this array defines the window size (also called kernel size or filter size). This design yields an architecture reminiscent of convolutional neural networks (CNNs), but importantly without weight sharing among nodes within a layer, thereby breaking translational symmetry. That aspect is actually close to the structure of visual processing layers in the biological brain, where weight sharing cannot be implemented. We refer to this architecture as locally coupled lattices (LCL) of a given window size [panel (b) of Fig.~\ref{fig:MNIST}].

    Alternatively, weight sharing can be imposed across nodes within the same layer, which restores translational symmetry. In this case, layers may also include multiple channels, producing an architecture closely resembling CNNs. We refer to this as a CNN-like architecture. Such models can be implemented physically in optical neural networks, for example through Fourier optics~\cite{miscuglio2020massively, fan2023experimental} or integrated photonics~\cite{bagherian2018onchipopticalconvolutionalneural}.  

    For completeness, we also consider the most basic form of layered networks: densely connected layered networks (DL). The difference of performance between DL and other comparable networks reveals the effect of architecture.

    During the training, we use the same nudging force as in Iris classification and have $\eta = 10^{-4}$ and $M_{init}=1$.

    We train networks of the three architectures on the full MNIST dataset. The structures of LCL, CNN-like, and DL networks are illustrated in panel (b) of Fig.~\ref{fig:MNIST}. Each network consists of a $28 \times 28$ input layer and a 10-node output layer which is interpreted as a probability distribution in the same way as for Iris. The hidden layers differ in design: LCL networks contain a single lattice layer with intra-layer couplings, CNN-like networks consist of multiple layers without intra-layer connections, and DL networks are one-dimensional arrays densely coupled to adjacent layers.  

    We first examine LCL networks, with results shown in panel (c) of Fig.~\ref{fig:MNIST}. Networks without intra-layer connections and with varying window sizes were trained for 200 epochs, using mini-batches of 200 randomly selected images in each epoch. All networks considered here include a single hidden layer. The effect of window size is illustrated in the upper graph of panel (c), compared against a baseline network without a hidden layer (blue solid line labeled “None”). All LCL networks outperform the baseline, with the $6 \times 6$ window achieving a test accuracy of 96.05\%. The middle graph shows the influence of stride size, analogous to the stride parameter in digital CNNs (allowing for resolution reduction). The results indicate that non-trivial strides, typically used for data compression, reduce network performance. The lower graph presents the effect of intra-layer connections. Contrary to the single-layer case, where more complex architectures improved performance, here an LCL network with 4NSQ intra-layer structure performs worse than simpler configurations such as SQ or unconnected hidden layers.  

    Panel (d) of Fig.~\ref{fig:MNIST} compares LCL networks with CNN-like and DL networks. In the left graph, CNN-like networks with $6 \times 6$ filters and varying channel numbers are trained for 300 epochs under the same protocol. The LCL network (black line) outperforms the 1-channel CNN-like network, matches the 2-channel case with greater stability, but is surpassed by CNN-like networks with 5 and 10 channels, which reach test accuracies of 97.64\% and 98.09\%, respectively. The right graph compares LCL and DL networks of the same number of trainable parameters. Error curves of LCL are shown as solid lines, while those of DL networks with the same number of parameters are plotted as dashed lines of the same color. LCL networks consistently achieve lower test error and greater stability than their DL counterparts.

    Panel (e) of Fig.~\ref{fig:MNIST} shows the influence of parameter count on network performance across different architectures. According to the result, CNN-like networks (labeled as ``CNN") performs the best with the same number of trainable parameters. LCL networks outperform DL networks with the same number of parameters. One can also see the negative effect of intralayer connections on the performance, which is mentioned above.


\section{Discussion}

    In this paper, we investigated the learning capabilities of networks with physically realizable architectures. We assessed how architectural features influence node responses and the evolution of couplings across lattices of varying complexity. Introducing longer skip connections enhances the network’s ability to maintain long-range responses during training, leading to improved performance. The networks also exhibit signs of self-organization: the lattice separates into an affected region, where node responses are strong and couplings are significantly strengthened, and a marginal region, where nodes show weak responses to input perturbations and couplings remain weak. Nevertheless, training lattices remains more challenging than training all-to-all networks; for instance, XOR can be successfully trained on an all-to-all network within 1,000 epochs \cite{wang2024training}, whereas a 4NSQ lattice requires approximately 20,000 epochs.

    Using the Iris dataset, we further explored the effect of depth and compared lattice performance to all-to-all and densely connected layered (DL) networks. For both adjacent and 1-separated input arrangements, lattice performance generally declines with increasing depth and remains below that of all-to-all networks. Surprisingly, lattices can still match or even surpass DL networks in some cases. This can be attributed to skip connections in specific lattice architectures, which provide cross-layer links that preserve information from previous layers, analogous to residual connections in deep neural networks \cite{he2016deep}. Identifying similar architectures in physical networks represents a promising direction for future work.

    Finally, we demonstrated the potential of layered networks by training LCL and CNN-like architectures on the full MNIST dataset. Both architectures achieve excellent performance, significantly surpassing linear classification in test accuracy. Compared to DL networks, LCL networks exhibit higher accuracy and greater stability, highlighting the advantages of local and skip connections in layered physical networks.

    Overall, these results suggest promising prospects for implementing neuromorphic computing using lattice models, with implications for more complex physical network architectures.

\section{Acknowledgment}

\bibliographystyle{apsrev4-2}
\bibliography{sample}

@article{scellier2017equilibrium,
  title={Equilibrium propagation: Bridging the gap between energy-based models and backpropagation},
  author={Scellier, Benjamin and Bengio, Yoshua},
  journal={Frontiers in computational neuroscience},
  volume={11},
  pages={24},
  year={2017},
  publisher={Frontiers Media SA}
}

@article{ernoult2020equilibrium,
  title={Equilibrium propagation with continual weight updates},
  author={Ernoult, Maxence and Grollier, Julie and Querlioz, Damien and Bengio, Yoshua and Scellier, Benjamin},
  journal={arXiv preprint arXiv:2005.04168},
  year={2020}
}

@article{scellier2022agnostic,
  title={Agnostic physics-driven deep learning},
  author={Scellier, Benjamin and Mishra, Siddhartha and Bengio, Yoshua and Ollivier, Yann},
  journal={arXiv preprint arXiv:2205.15021},
  year={2022}
}

@article{falk2025temporal,
  title={Temporal Contrastive Learning through implicit non-equilibrium memory},
  author={Falk, Martin J and Strupp, Adam T and Scellier, Benjamin and Murugan, Arvind},
  journal={Nature Communications},
  volume={16},
  number={1},
  pages={2163},
  year={2025},
  publisher={Nature Publishing Group UK London}
}

@article{kendall2020training,
  title={Training end-to-end analog neural networks with equilibrium propagation},
  author={Kendall, Jack and Pantone, Ross and Manickavasagam, Kalpana and Bengio, Yoshua and Scellier, Benjamin},
  journal={arXiv preprint arXiv:2006.01981},
  year={2020}
}

@article{wright2022deep,
  title={Deep physical neural networks trained with backpropagation},
  author={Wright, Logan G and Onodera, Tatsuhiro and Stein, Martin M and Wang, Tianyu and Schachter, Darren T and Hu, Zoey and McMahon, Peter L},
  journal={Nature},
  volume={601},
  number={7894},
  pages={549--555},
  year={2022},
  publisher={Nature Publishing Group UK London}
}

@article{wang2024training,
  title={Training coupled phase oscillators as a neuromorphic platform using equilibrium propagation},
  author={Wang, Qingshan and Wanjura, Clara C and Marquardt, Florian},
  journal={Neuromorphic Computing and Engineering},
  volume={4},
  number={3},
  pages={034014},
  year={2024},
  publisher={IOP Publishing}
}

@article{laborieux2021scaling,
  title={Scaling equilibrium propagation to deep convnets by drastically reducing its gradient estimator bias},
  author={Laborieux, Axel and Ernoult, Maxence and Scellier, Benjamin and Bengio, Yoshua and Grollier, Julie and Querlioz, Damien},
  journal={Frontiers in neuroscience},
  volume={15},
  pages={633674},
  year={2021},
  publisher={Frontiers Media SA}
}

@article{dillavou2022demonstration,
  title={Demonstration of decentralized physics-driven learning},
  author={Dillavou, Sam and Stern, Menachem and Liu, Andrea J and Durian, Douglas J},
  journal={Physical Review Applied},
  volume={18},
  number={1},
  pages={014040},
  year={2022},
  publisher={APS}
}

@misc{bagherian2018onchipopticalconvolutionalneural,
      title={On-Chip Optical Convolutional Neural Networks}, 
      author={Hengameh Bagherian and Scott Skirlo and Yichen Shen and Huaiyu Meng and Vladimir Ceperic and Marin Soljacic},
      year={2018},
      eprint={1808.03303},
      archivePrefix={arXiv},
      primaryClass={cs.ET},
      url={https://arxiv.org/abs/1808.03303}, 
}

@article{miscuglio2020massively,
  title={Massively parallel amplitude-only Fourier neural network},
  author={Miscuglio, Mario and Hu, Zibo and Li, Shurui and George, Jonathan K and Capanna, Roberto and Dalir, Hamed and Bardet, Philippe M and Gupta, Puneet and Sorger, Volker J},
  journal={Optica},
  volume={7},
  number={12},
  pages={1812--1819},
  year={2020},
  publisher={Optical Society of America}
}

@article{fan2023experimental,
  title={Experimental realization of convolution processing in photonic synthetic frequency dimensions},
  author={Fan, Lingling and Wang, Kai and Wang, Heming and Dutt, Avik and Fan, Shanhui},
  journal={Science Advances},
  volume={9},
  number={32},
  pages={eadi4956},
  year={2023},
  publisher={American Association for the Advancement of Science}
}

@inproceedings{he2016deep,
  title={Deep residual learning for image recognition},
  author={He, Kaiming and Zhang, Xiangyu and Ren, Shaoqing and Sun, Jian},
  booktitle={Proceedings of the IEEE conference on computer vision and pattern recognition},
  pages={770--778},
  year={2016}
}

@article{laydevant2024training,
  title={Training an ising machine with equilibrium propagation},
  author={Laydevant, J{\'e}r{\'e}mie and Markovi{\'c}, Danijela and Grollier, Julie},
  journal={Nature Communications},
  volume={15},
  number={1},
  pages={3671},
  year={2024},
  publisher={Nature Publishing Group UK London}
}

@article{rageau2025training,
  title={Training and synchronizing oscillator networks with Equilibrium Propagation},
  author={Rageau, Theophile and Grollier, Julie},
  journal={Neuromorphic Computing and Engineering},
  year={2025}
}

@article{stern2021supervised,
  title={Supervised learning in physical networks: From machine learning to learning machines},
  author={Stern, Menachem and Hexner, Daniel and Rocks, Jason W and Liu, Andrea J},
  journal={Physical Review X},
  volume={11},
  number={2},
  pages={021045},
  year={2021},
  publisher={APS}
}

@article{lopez2023self,
  title={Self-learning machines based on Hamiltonian echo backpropagation},
  author={Lopez-Pastor, Victor and Marquardt, Florian},
  journal={Physical Review X},
  volume={13},
  number={3},
  pages={031020},
  year={2023},
  publisher={APS}
}

@article{wanjura2025quantum,
  title={Quantum equilibrium propagation for efficient training of quantum systems based on Onsager reciprocity},
  author={Wanjura, Clara C and Marquardt, Florian},
  journal={Nature Communications},
  volume={16},
  number={1},
  pages={6595},
  year={2025},
  publisher={Nature Publishing Group UK London}
}

@article{pourcel2025lagrangian,
  title={Lagrangian-based Equilibrium Propagation: generalisation to arbitrary boundary conditions \& equivalence with Hamiltonian Echo Learning},
  author={Pourcel, Guillaume and Basu, Debabrota and Ernoult, Maxence and Gilra, Aditya},
  journal={arXiv preprint arXiv:2506.06248},
  year={2025}
}

@article{massar2025equilibrium,
  title={Equilibrium propagation for learning in Lagrangian dynamical systems},
  author={Massar, Serge},
  journal={Physical Review E},
  volume={112},
  number={3},
  pages={035304},
  year={2025},
  publisher={APS}
}

@article{de2025learning,
  title={Learning in a Multifield Coherent Ising Machine},
  author={de Bos, Daan and Serra-Garcia, Marc},
  journal={arXiv preprint arXiv:2502.12020},
  year={2025}
}

@article{wetzstein2020inference,
  title={Inference in artificial intelligence with deep optics and photonics},
  author={Wetzstein, Gordon and Ozcan, Aydogan and Gigan, Sylvain and Fan, Shanhui and Englund, Dirk and Solja{\v{c}}i{\'c}, Marin and Denz, Cornelia and Miller, David AB and Psaltis, Demetri},
  journal={Nature},
  volume={588},
  number={7836},
  pages={39--47},
  year={2020},
  publisher={Nature Publishing Group UK London}
}

@article{hoppensteadt1999oscillatory,
  title={Oscillatory neurocomputers with dynamic connectivity},
  author={Hoppensteadt, Frank C and Izhikevich, Eugene M},
  journal={Physical Review Letters},
  volume={82},
  number={14},
  pages={2983},
  year={1999},
  publisher={APS}
}

@article{scellier2018generalization,
  title={Generalization of equilibrium propagation to vector field dynamics},
  author={Scellier, Benjamin and Goyal, Anirudh and Binas, Jonathan and Mesnard, Thomas and Bengio, Yoshua},
  journal={arXiv preprint arXiv:1808.04873},
  year={2018}
}

@article{pai2023experimentally,
  title={Experimentally realized in situ backpropagation for deep learning in photonic neural networks},
  author={Pai, Sunil and Sun, Zhanghao and Hughes, Tyler W and Park, Taewon and Bartlett, Ben and Williamson, Ian AD and Minkov, Momchil and Milanizadeh, Maziyar and Abebe, Nathnael and Morichetti, Francesco and others},
  journal={Science},
  volume={380},
  number={6643},
  pages={398--404},
  year={2023},
  publisher={American Association for the Advancement of Science}
}

@article{wanjura2024fully,
  title={Fully nonlinear neuromorphic computing with linear wave scattering},
  author={Wanjura, Clara C and Marquardt, Florian},
  journal={Nature Physics},
  volume={20},
  number={9},
  pages={1434--1440},
  year={2024},
  publisher={Nature Publishing Group UK London}
}

@article{scellier2024quantum,
  title={Quantum equilibrium propagation: Gradient-descent training of quantum systems},
  author={Scellier, Benjamin},
  journal={arXiv preprint arXiv:2406.00879},
  year={2024}
}

@article{massar2025quantum,
  title={Equilibrium propagation: the quantum and the thermal cases},
  author={Massar, Serge and Mognetti, Bortolo Matteo},
  journal={Quantum Studies: Mathematics and Foundations},
  volume={12},
  number={1},
  pages={6},
  year={2025},
  publisher={Springer}
}

@article{hopfield1982neural,
  title={Neural networks and physical systems with emergent collective computational abilities.},
  author={Hopfield, John J},
  journal={Proceedings of the national academy of sciences},
  volume={79},
  number={8},
  pages={2554--2558},
  year={1982}
}

@article{farhat1985optical,
  title={Optical implementation of the Hopfield model},
  author={Farhat, Nabil H and Psaltis, Demetri and Prata, Aluizio and Paek, Eung},
  journal={Applied optics},
  volume={24},
  number={10},
  pages={1469--1475},
  year={1985},
  publisher={Optical Society of America}
}

@article{ackley1985learning,
  title={A learning algorithm for Boltzmann machines},
  author={Ackley, David H and Hinton, Geoffrey E and Sejnowski, Terrence J},
  journal={Cognitive science},
  volume={9},
  number={1},
  pages={147--169},
  year={1985},
  publisher={Elsevier}
}

@inproceedings{kim2009highly,
  title={A highly scalable restricted Boltzmann machine FPGA implementation},
  author={Kim, Sang Kyun and McAfee, Lawrence C and McMahon, Peter L and Olukotun, Kunle},
  booktitle={2009 International Conference on Field Programmable Logic and Applications},
  pages={367--372},
  year={2009},
  organization={IEEE}
}

@article{ticknor1987optical,
  title={Optical implementations in Boltzmann machines},
  author={Ticknor, Anthony J and Barrett, Harrison H},
  journal={Optical Engineering},
  volume={26},
  number={1},
  pages={16--21},
  year={1987},
  publisher={SPIE}
}

@article{li2018review,
  title={Review of memristor devices in neuromorphic computing: materials sciences and device challenges},
  author={Li, Yibo and Wang, Zhongrui and Midya, Rivu and Xia, Qiangfei and Yang, J Joshua},
  journal={Journal of Physics D: Applied Physics},
  volume={51},
  number={50},
  pages={503002},
  year={2018},
  publisher={IOP Publishing}
}

@article{oguz2023forward,
  title={Forward--forward training of an optical neural network},
  author={Oguz, Ilker and Ke, Junjie and Weng, Qifei and Yang, Feng and Yildirim, Mustafa and Dinc, Niyazi Ulas and Hsieh, Jih-Liang and Moser, Christophe and Psaltis, Demetri},
  journal={Optics Letters},
  volume={48},
  number={20},
  pages={5249--5252},
  year={2023},
  publisher={Optica Publishing Group}
}

@article{momeni2023backpropagation,
  title={Backpropagation-free training of deep physical neural networks},
  author={Momeni, Ali and Rahmani, Babak and Mall{\'e}jac, Matthieu and Del Hougne, Philipp and Fleury, Romain},
  journal={Science},
  volume={382},
  number={6676},
  pages={1297--1303},
  year={2023},
  publisher={American Association for the Advancement of Science}
}

@article{momeni2025training,
  title={Training of physical neural networks},
  author={Momeni, Ali and Rahmani, Babak and Scellier, Benjamin and Wright, Logan G and McMahon, Peter L and Wanjura, Clara C and Li, Yuhang and Skalli, Anas and Berloff, Natalia G and Onodera, Tatsuhiro and others},
  journal={Nature},
  volume={645},
  number={8079},
  pages={53--61},
  year={2025},
  publisher={Nature Publishing Group UK London}
}

@article{cin2025training,
  title={Training nonlinear optical neural networks with Scattering Backpropagation},
  author={Cin, Nicola Dal and Marquardt, Florian and Wanjura, Clara C},
  journal={arXiv preprint arXiv:2508.11750},
  year={2025}
}

@article{wagner1987multilayer,
  title={Multilayer optical learning networks},
  author={Wagner, Kelvin and Psaltis, Demetri},
  journal={Applied Optics},
  volume={26},
  number={23},
  pages={5061--5076},
  year={1987},
  publisher={Optical Society of America}
}

@article{spall2025training,
  title={Training neural networks with end-to-end optical backpropagation},
  author={Spall, James and Guo, Xianxin and Lvovsky, Alexander I},
  journal={Advanced Photonics},
  volume={7},
  number={1},
  pages={016004--016004},
  year={2025},
  publisher={Society of Photo-Optical Instrumentation Engineers}
}

@article{stern2024training,
  title={Training self-learning circuits for power-efficient solutions},
  author={Stern, Menachem and Dillavou, Sam and Jayaraman, Dinesh and Durian, Douglas J and Liu, Andrea J},
  journal={APL Machine Learning},
  volume={2},
  number={1},
  year={2024},
  publisher={AIP Publishing}
}

@article{wycoff2022desynchronous,
  title={Desynchronous learning in a physics-driven learning network},
  author={Wycoff, Jacob F and Dillavou, Sam and Stern, Menachem and Liu, Andrea J and Durian, Douglas J},
  journal={The Journal of Chemical Physics},
  volume={156},
  number={14},
  year={2022},
  publisher={AIP Publishing}
}

\appendix

\end{document}